%% file: main.tex
\documentclass[sigconf]{acmart}

\usepackage{booktabs} 
\setcopyright{rightsretained}
\usepackage{epigraph}

\renewcommand\footnotetextcopyrightpermission[1]{} 

\acmDOI{}

\acmISBN{}

\acmConference[ML4Creativity'17]{Workshop on Machine Learning for Creativity, SIGKDD}{August 2017}{Nova Scotia, Canada} 
\acmYear{2017}
\copyrightyear{2017}

\acmPrice{}

\graphicspath{{./}}
\begin{document}
\title{mAnI: Movie Amalgamation using Neural Imitation}
\subtitle{Visualizing the Movie while Reading a Book}

\author{Naveen Panwar, Shreya Khare, Neelamadhav Gantayat, Rahul Aralikatte, Senthil Mani, Anush Sankaran}
\affiliation{%
  \institution{IBM Research, India}
}
\email{{naveen.panwar, shkhare4, neelamadhav, rahul.a.r, sentmani, anussank}@in.ibm.com}

\renewcommand{\shortauthors}{Panwar et al.}

\begin{abstract}
Cross-modal data retrieval has been the basis of various creative tasks performed by Artificial Intelligence (AI). One such highly challenging task for AI is to convert a book into its corresponding movie, which most of the creative film makers do as of today. In this research, we take the first step towards it by visualizing the content of a book using its corresponding movie visuals. Given a set of sentences from a book or even a fan-fiction written in the same universe, we employ deep learning models to visualize the input by stitching together relevant frames from the movie. We studied and compared three different types of setting to match the book with the movie content: (i) Dialog model: using only the dialog from the movie, (ii) Visual model: using only the visual content from the movie, and (iii) Hybrid model: using the dialog and the visual content from the movie. Experiments on the publicly available MovieBook dataset shows the effectiveness of the proposed models.
\end{abstract}

%
%



\maketitle

\input{1_introduction}
\input{2_Literature}

\input{2_Dataset}

\input{3_Proposed_Approach}
\input{4_Experimental_Study}
\input{7_Conclusion}

\bibliographystyle{ACM-Reference-Format}
\bibliography{sigproc} 

\end{document}

%% file: 1_introduction.tex
\section{Introduction}
Being able to fluently understand, retrieve, and generate cross-modal data, like humans do, has been the holy grail search in Artificial Intelligence (AI). Language and vision has been considered as the most common and challenging domains to measure the growth of artificial intelligence. Describing an image in words (image captioning) and imagining a text through images (visual abstraction/ description) is highly natural and seamless for human beings. While reading a gripping novel or a book, we often tend imagine the storyline and the plots through visuals. If a corresponding movie or video exists for a book, most of the imaginative visuals are borrowed from the movie and mapped with the book stories. Another common example is of the movie director (or a movie creation crew), who produces a movie from a book or storyline through creative visualizations. 

Consider the example book snippet from \textit{Harry Potter and the Philosopher's Stone} - 
\epigraph{Professor McGonagall peered sternly over her glasses at Harry. \\ "I want to hear you're training hard, Potter, or I may change my mind about punishing you." \\
Then she suddenly smiled. \\
"Your father would have been proud," she said.
"He was an excellent Quidditch player himself."}{Prof. McGonagall}

\begin{figure*}[ht]
	\centering
	\includegraphics[width=6.8in]{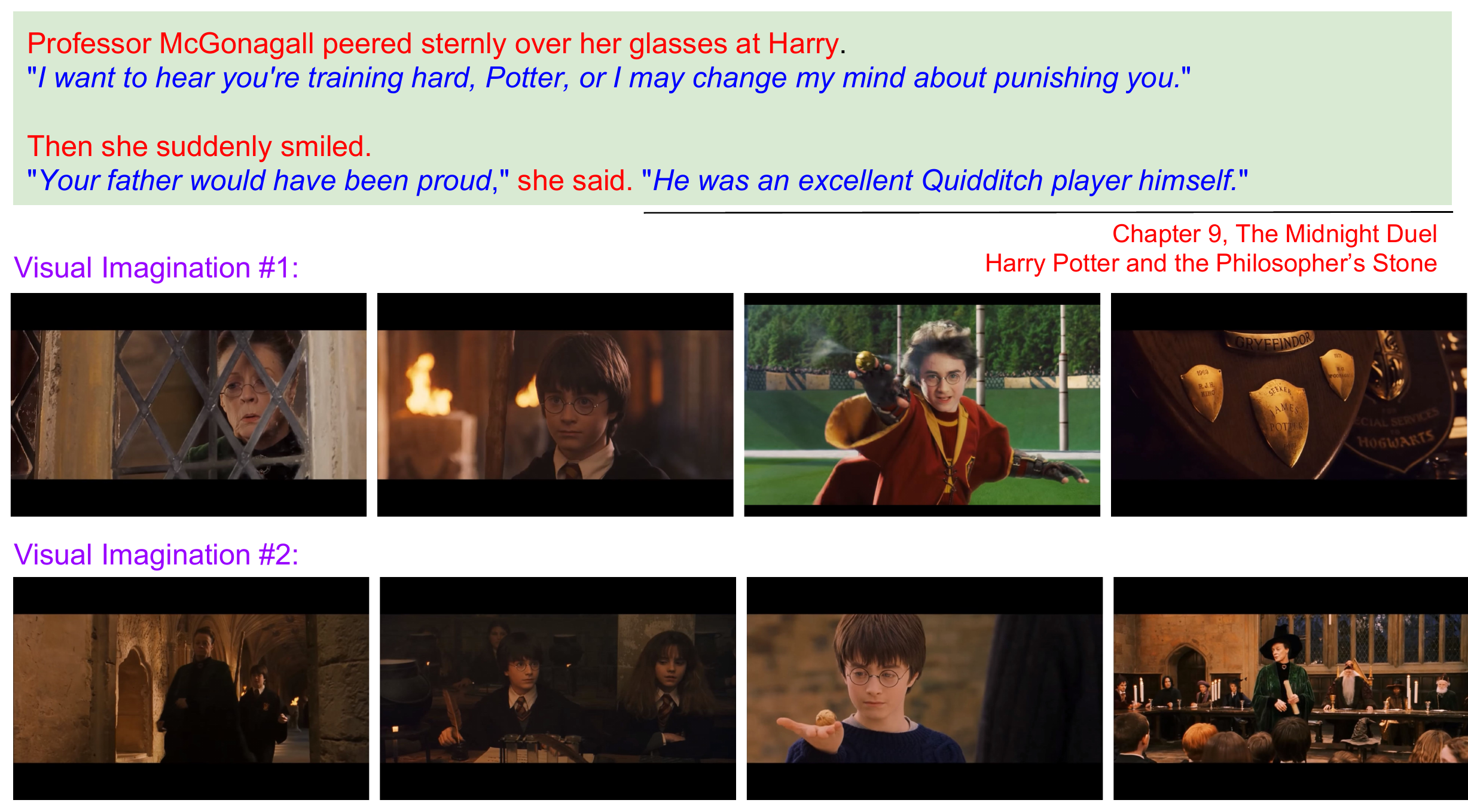}
	\caption{Visually imagining a story snippet from the \textit{Harry Potter and the Philosopher's Stone} book through its corresponding movie visuals. The visuals are obtained from the MovieBook dataset~\cite{zhu2015aligning}.}
	\label{fig:motivation_1}
\end{figure*}

\noindent It is natural for readers to imagine and visualize this book snippet through snippets from the corresponding movie. Figure~\ref{fig:motivation_1} shows two possible visualizations of the book snippet as imagined by two different readers. These visualizations provide information rich interpretations to the books. The examples are not only restricted to the actual book snippets but also towards any fandom, as the following one - 

\epigraph{"As such I do not expect anyone to understand the subtleties of using machine learning in creativity", scowled Snape, as he charged into the dark classroom and glared into Harry's pale blue eyes.\\
"However, our celebrity Harry Potter," he paused, "could probably enlighten us with what would happen if I added a LSTM over a CNN"}{Fan Fiction}

\begin{figure*}[!h]
	\centering
	\includegraphics[width=6.8in]{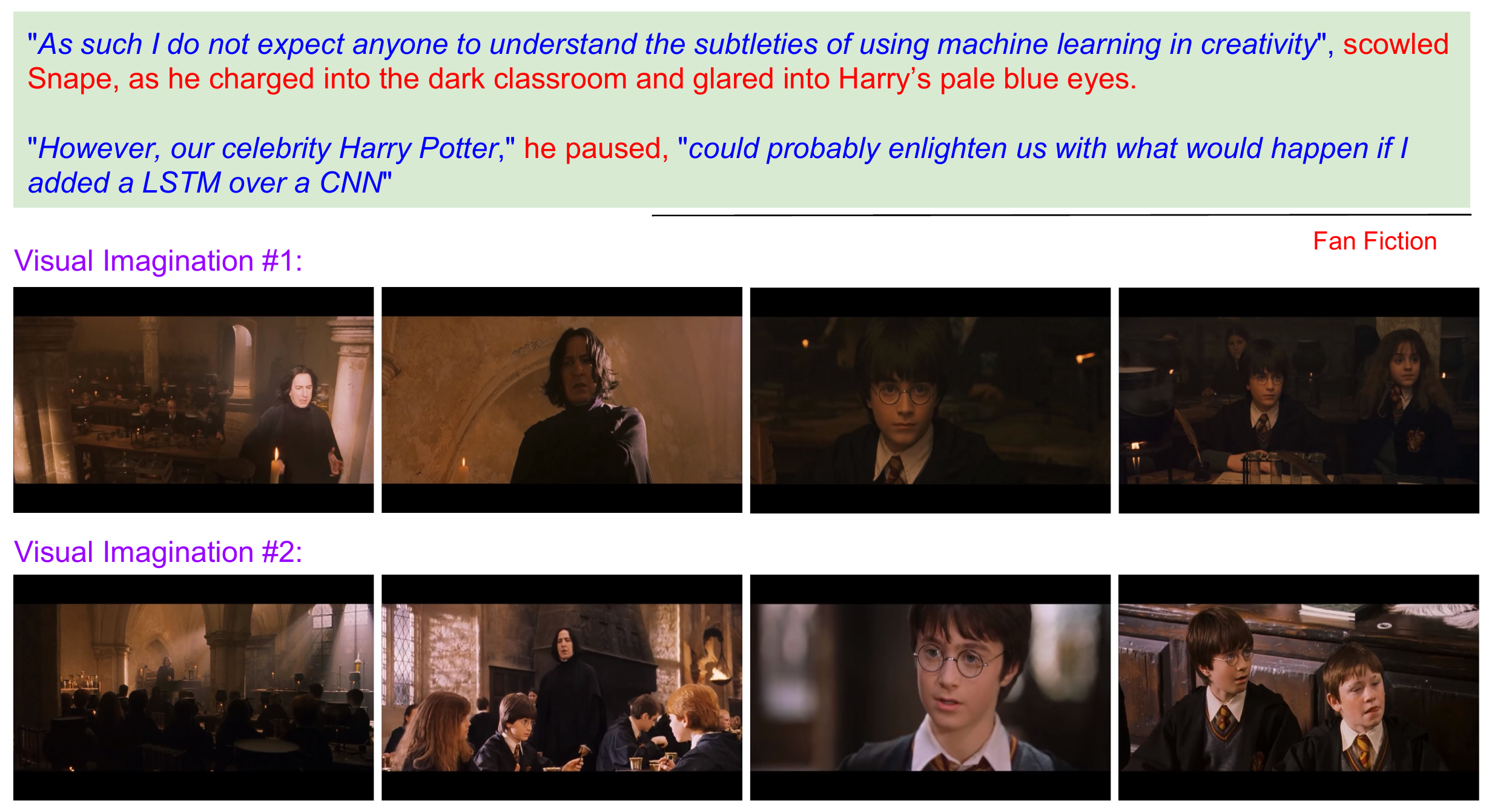}
	\caption{Visually imagining a random fan fiction story snippet from the \textit{Harry Potter} fandom through the corresponding movie visuals. The visuals are obtained from the MovieBook dataset~\cite{zhu2015aligning}.}
	\label{fig:motivation_2}
\end{figure*}

\noindent Figure~\ref{fig:motivation_2} shows two possible visualizations of the fan fiction of the same book, \textit{Harry Potter and the Philosopher's Stone}. Motivated by such human behaviors, in this research, we attempt to describe constituent parts of a book or a story through its corresponding movie visuals.

\begin{table*}[tb]
	\scriptsize
	\begin{tabular}{|p {90pt}|p {20pt}|p {25pt}|p {30pt}|p {25pt}|p {33pt}|p {20pt}|p {20pt}|p {25pt}|p {20pt}|p {20pt}|}\hline
		&\multicolumn{6}{c|}{\textbf{BOOK}}&\multicolumn{2}{c|}{\textbf{MOVIE}} & \multicolumn{2}{c|}{\textbf{ANNOTATION}}\\ \hline
		Title  &   \# sent.  &   \# words  &   \# unique words avg.  &   \# words per sent.  &  max \# words per sent.  &  \# paragraphs  &   \# shots  &   \# sent. in subtitles  &  \# dialog align.  &   \# visual align.  \\ \hline
		Gone Girl  &  12,603  &  1,48,340  &  3,849  &  15  &  153  &  3,927  &  2,604  &  2,555  &  76  &  106  \\
		Fight Club  &  4,229  &  48,946  &  1,833  &  14  &  90  &  2,082  &  2,365  &  1,864  &  104  &  42  \\
		No Country for Old Men  &  8,050  &  69,824  &  1,704  &  10  &  68  &  3,189  &  1,348  &  889  &  223  &  47  \\
		Harry Potter and the Sorcerers Stone  &  6,458  &  78,596  &  2,363  &  15  &  227  &  2,925  &  2,647  &  1,227  &  164  &  73  \\
		Shawshank Redemption  &  2,562  &  40,140  &  1,360  &  18  &  115  &  637  &  1,252  &  1,879  &  44  &  12  \\
		The Green Mile  &  9,467  &  1,33,241  &  3,043  &  17  &  119  &  2,760  &  2,350  &  1,846  &  208  &  102  \\
		American Psycho  &  11,992  &  1,43,631  &  4,632  &  16  &  422  &  3,945  &  1,012  &  1,311  &  278  &  85  \\
		One Flew Over the Cuckoo Nest  &  7,103  &  1,12,978  &  2,949  &  19  &  192  &  2,236  &  1,671  &  1,553  &  64  &  25  \\
		The Firm  &  15,498  &  1,35,529  &  3,685  &  11  &  85  &  5,223  &  2,423  &  1,775  &  82  &  60  \\
		Brokeback Mountain  &  638  &  10,640  &  470  &  20  &  173  &  167  &  1,205  &  1,228  &  80  &  20  \\
		The Road  &  6,638  &  58,793  &  1,580  &  10  &  74  &  2,345  &  1,108  &  782  &  126  &  49  \\\hline
		All  &  85,238  &  9,80,658  &  9,032  &  15  &  156  &  29,436  &  19,985  &  16,909  &  1,449  &  621  \\ \hline
	\end{tabular}
	\caption{\label{table1} Statistics for MovieBook dataset~\cite{zhu2015aligning} with ground-truth for alignment between books and their movie releases.}
\end{table*}
The publicly available MovieBook~\cite{zhu2015aligning} dataset contains manually defined alignment of $11$ movies with their corresponding books. Given a book snippet, we retrieve a sequence of movie snippets describing that book snippet, using three independent models:
\begin{enumerate}
	\item \textbf{Dialog model}: Relevant movie snippets are retrieved by matching the text dialog of the movie with the input book snippet using a skip-thought model~\cite{kiros2015skip}
	\item \textbf{Visual model}: Relevant movie snippets are retrieved by matching only the visual cues of the movie scene with the input book snippet using a neural-storyteller model~\footnote{https://github.com/ryankiros/neural-storyteller}
	\item \textbf{Hybrid model}: Relevant movie snippets are retrieved using both the text dialog and the visual cues from the movie scene
\end{enumerate}

The rest of the paper is organized as follows; Section 2 talks about existing literature, Section 3 details the dataset used in this research, Section 3 explains the technical details of the proposed approach, Section 5 discusses the experimental results, and Section 6 concludes with some future directions.

%% file: 2_Literature.tex
\section{Literature Study}
The defined problem statement requires the understanding of both the domains: video analysis and natural langauge processing. Textual content and concept based video retrieval has been well explored in the literature~\cite{sivic2003video, campbell2007ibm, lew2006content}. Yang and Meinel~\cite{yang2014content} used Optical Character Recognition (OCR) and Automatic Speech Recognition (ASR) to transcribe content from video lectures and perform querying over the extracted content. Tian et al.~\cite{tian2017unified} further extended this by tracking textual content across the frames in a video for better content generation. Xu et al.~\cite{xu2015jointly} learnt a joint text-video embedding model built over independently learnt deep models of language semantic understanding and video embedding. Thus, they were able to perform both video to text generation and text based video retrieval using the joint model.  

Ng et al.~\cite{yue2015beyond} considered each frame of a video as a word in a sentence and learnt an LSTM network to temporally embed the video. The representation for each frame was obtained using a deep CNN making the overall network as a CNN-LSTM deep network. Donahue et al.~\cite{donahue2015long} proposed a Long-Term Recurrent Convolutional Network (LRCN) model for conditionally embedding the video based on the task to be performed. Venugopalan et al.~\cite{venugopalan2015sequence} learnt a sequence to sequence model to encode a video frame sequence using an LSTM network and decode its corresponding caption using a conditional LSTM. For understanding large pieces of text, Le and Mikolov~\cite{le2014distributed} extended a word representation \textit{word2vec} to learn paragraph and document level representation. Arora et al.~\cite{arora2016simple} proposed a simple method of averaging the word embeddings over a sentence and modifying it using PCA. Recently, Kiros et al.~\cite{kiros2015skip} came up with an unsupervised method of learning sentence representation called skip-thought vectors which provided comparable results in $8$ different tasks without the need for task adaptation.

One of the closest work to our research is the MovieQA system proposed by Tapaswi et al.~\cite{tapaswi2016movieqa}. A memory network based question answering system is built on a movie corpus using multiple sources of information such as movie plot, movie video, subtitle, scripts, and Described Video Service (DVS) transcriptions. Our work considers the book content as the input which is, in general, more prose and descriptive than a movie plot or script. Tapaswi et al.~\cite{tapaswi2015book2movie} further proposed \textit{Book2Movie} which aims to align book chapters to its corresponding movie scenes. We are working at a much granular level of sentences rather than an entire chapter, which is a more challenging task. Our work is built upon Zhu et al.~\cite{zhu2015aligning} aligning books and movies at sentence level. While they have performed experiments on describing movies in terms of the book, we attempt to describe the book in terms of the movie which is considered a much more creative problem.

%% file: 2_Dataset.tex
\section{Dataset}

Built upon the work by Zhu et al.~\cite{zhu2015aligning}, the MovieBook dataset is the highly relevant for our problem statement. The dataset contains visual clips (roughly spanning for few seconds) from movies, corresponding dialogue text (SRT) for the visual clips, and small chunks of book text (roughly 3-10 lines) for $11$ different books. A manual alignment is available for a part of each book and each alignment is done using one of the three cues: (i) Visual cue based on the movie clip, (ii) Dialog cue based on the dialog spoken during that clip, and (iii) Audio cue based on the audio during that clip. The properties of this dataset are shown in Table~\ref{table1}. From the collection of 11 book-movie pairs, there are a total of $29,436$ book paragraphs, $19,985$ movie shots, and $16,909$ sentences in dialog subtitles. Using this corpus, a total of $1,449$ (book paragraph, movie shot) pairs were manually aligned using the dialog subtitles while $621$ pairs were aligned using the visual content of the movie shot.

Additional to the MovieBook dataset, a huge corpus of books is used to train a model for sentence representation. The BookCorpus dataset has more than $11,000$ from $16$ different genres containing more than $11$ million sentences and a skip-thought model~\cite{kiros2015skip} is trained to learn a sentence representation. The pretrained model is already publicly available at: \url{https://github.com/ryankiros/skip-thoughts}.

%% file: 3_Proposed_Approach.tex
\section{Proposed Approach}

\begin{figure*}[!h]
	\centering
	\includegraphics[width=6.8in]{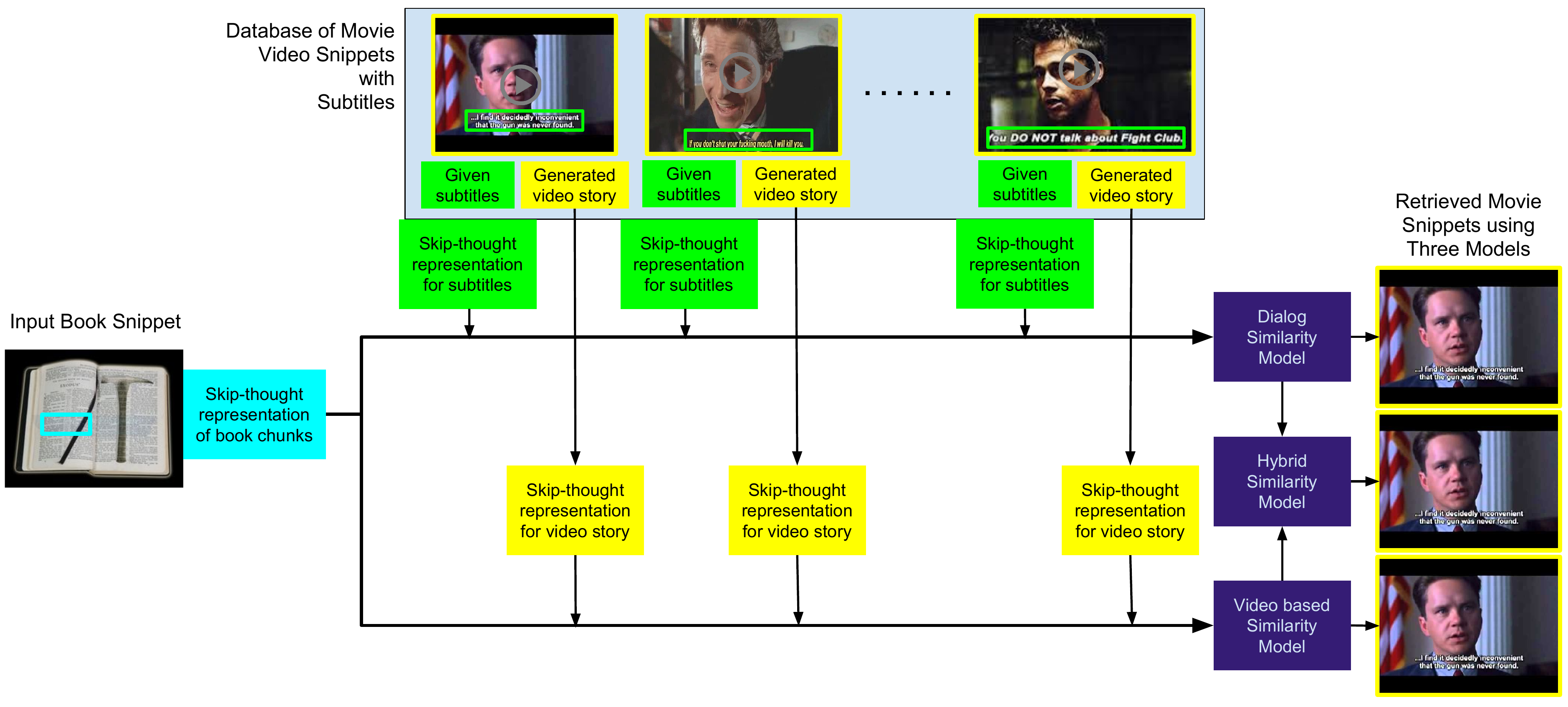}
	\caption{Illustration of the proposed approach to describe book snippets in terms of its movie video snippets.}
	\label{fig:approach}
\end{figure*}

The overall proposed approach has three different models and is illustrated in Figure~\ref{fig:approach}. The individual steps and their training procedure is explained in detail in this section.

\begin{figure*}[!h]
	\centering
	\includegraphics[width=6.7in]{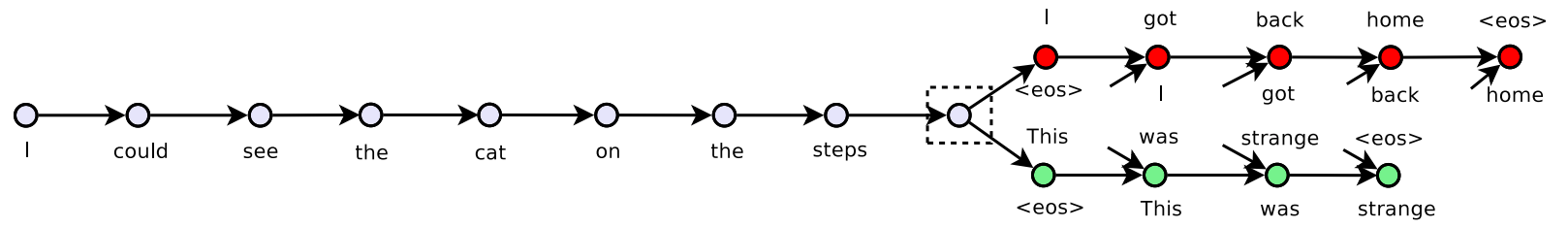}
	\caption{The encoder-decoder based skip-thought model design, proposed by Kiros et al~\cite{kiros2015skip}. This skip-thought model pretrained on the BookCorpus dataset is used to extract sentence representations of both the book paragraphs and the movie dialogs.}
	\label{fig:skipthought}
\end{figure*}

\subsection{Book Sentence and Movie Dialog Representation}

For every chunk of the book or a dialog snippet, a sentence representation model is learnt using skip-thought vectors~\cite{kiros2015skip}, one of the state-of-art models for unsupervised learning of text sequences. The skip-thought vector model is a natural encoder-decoder style extension of skip-gram model for word embedding learning. Given a tuple of three sentences, \{$s_{i-1}$, $s_i$, $s_{i+1}$\}, the a RNN model encodes the sentence $s_i$ and two decoders attempts to predict the sentence $s_{i-1}$ and $s_{i+1}$, conditional on the encoding, as shown in Figure~\ref{fig:skipthought}. Thus, such a model requires tuples of three sentences and can be trained in an unsupervised fashion. Kiros et al.~\cite{kiros2015skip} further show that a generic sentence representation model trained on a huge corpus of books can be directly used in eight different applications without the need for fine-tuning or task adaption. Owing to the generalizable nature of the skip-thought model, we use the available pre-trained model for directly extracting the representation of both book sentences and movie dialogue.

\begin{figure*}[!h]
	\centering
	\includegraphics[width=6.8in]{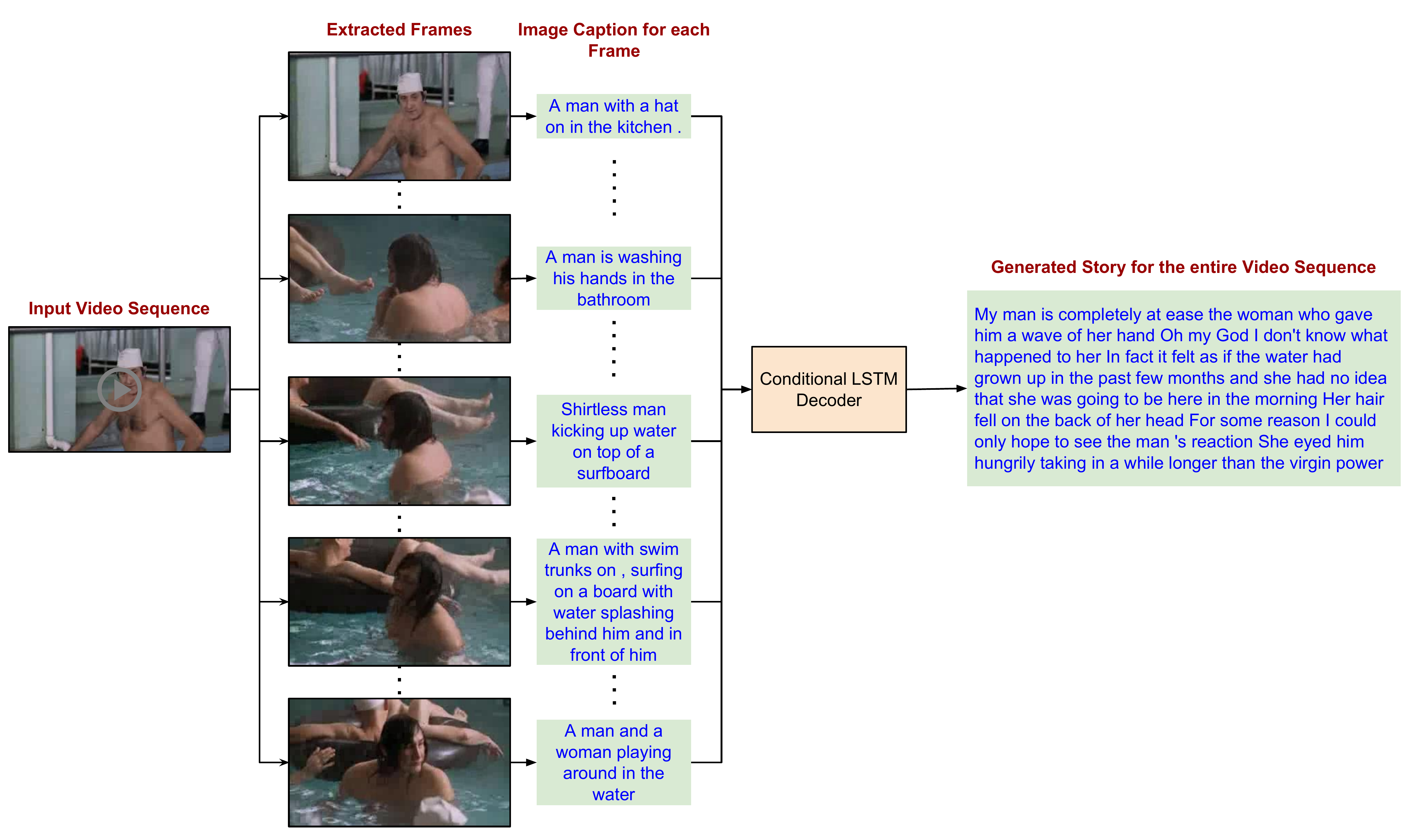}
	\caption{An illustration from the movie, American Pyscho, explaining the proposed approach for generating a textual story from a video.}
	\label{fig:video}
\end{figure*}

\subsection{Movie Video Representation}
\label{video}
Representing videos as an embedded vector representation is well studied. In this research, we want to textually  describe a movie clip so that semantic similarity could be computed with the book snippet. Image captioning and video captioning techniques could generate a single sentence caption for an image or a video. Recently, neural-storyteller~\footnote{\url{https://github.com/ryankiros/neural-storyteller}} conditioned the image caption on an RNN to generate a longer story to explain a single image. In this research, we explain the neural-storyteller style of model to generate a longer story for a video than for a single image. Given a video frame, an image caption is generated for every frame using an encoder-decoder model as proposed in~\cite{tapaswi2015book2movie}. Conditioned on the combined frame captions, an RNN decoder generates a story explaining the entire video clip. The details of the model is explained here: \url{https://github.com/ryankiros/neural-storyteller} and the process is illustrated in Figure~\ref{fig:video}. From an input video sequence, frame are sampled are regular intervals at 2fps. For every frame, a caption is generated using a standard image captioning model. All the generated captions are pooled and provided as input to a conditional LSTM decoder, which generates story that represents the entire video sequence and not each frame in the video. It can be observed from the generated story shown in Figure~\ref{fig:video}, that the swimming pool is semantically being mapped to water, the person is being mapped as ``her" due to the presence of long hair, and specific semantic attributes are extracted such as shirtless man, being top of surfboard. These semantically extracted text could be used to map with the book paragraph which are typically descriptive, in nature.

\subsection{Extraction through Dialog Model}
\label{model1}
In this model, a similarity metric learnt between the book sentences and only the dialog (SRT) in the video, without leveraging any visual content of the movie. Given a pair of book sentence and dialog, their respective representations, $\vec{b}$ and $\vec{d}$ are computed using the skip-thought model. As proposed in~\cite{TaiSM15}, $\vec{b}.\vec{d}$ and $abs(\vec{b} - \vec{d})$ are computed and concatenated. Over these representations, a regression based semantic similarity model is trained~\cite{TaiSM15}. Here the regression model is binary,  predicting the input pair as \{match, non-match\}.

During test phase, for a given book sentence or a random fan fiction sentence, its skip-thought representation is calculated and the semantic relevance is computed against all the dialog sentences available. A list of those dialog sentences above a threshold, $t$, is shortlisted as the relevant movie parts that explains the input book sentence. The video clips corresponding to the retrieved dialog sentences are stitched together and provided to the user.

\subsection{Extraction through Visual Model}
\label{model2}
For this model, only the book sentences and the video clips are used while the dialog sentences are not used. For a given video clip, a story explaining that video clip is automatically generated using the approach proposed in the Section~\ref{video}. For the automatically extracted story, a skip-thought representation is extracted, so that, both the book sentence and video clips is in the same feature space. In this space, the similarity classifier can be trained and tested in the similar way, as explained in Section~\ref{model1}.

\subsection{Extraction through Hybrid Model}
To match a book with the corresponding video clip, in this hybrid model, we leverage both the video information as well as the dialog information. For a given book sentence, the similarity score for all the dialog sentences is obtained using the dialog model explained in Section~\ref{model1} and the similarity score is obtained with all the video clips using the Visual model explained in Section~\ref{model2}. A sum score fusion is performed between the two lists of obtained similarity score, and a the threshold is applied on the fused score. The movie clips corresponding to the retrievals are stitched to provide the book visualization.

%% file: 4_Experimental_Study.tex
\section{Experimental Study}

The experiments are conducted on the publicly available MovieBook Corpus. The only trainable model is the semantic similarity model explained in Section~\ref{model1}. The entire data is split between $60\%$ for training, $20\%$ for validation, and $20\%$ testing. Thus, for the Dialog model, there are $1842$ for training samples and $616$ test samples while for the Visual model $776$ samples are used for training and $258$ samples for testing. To compare with the proposed similarity model, a cosine distance based similarity metric as well the similarity model proposed by Tai et al.~\cite{TaiSM15} and trained on SICK dataset, are used.

\begin{figure}[!h]
	\centering
	\includegraphics[width=3.4in]{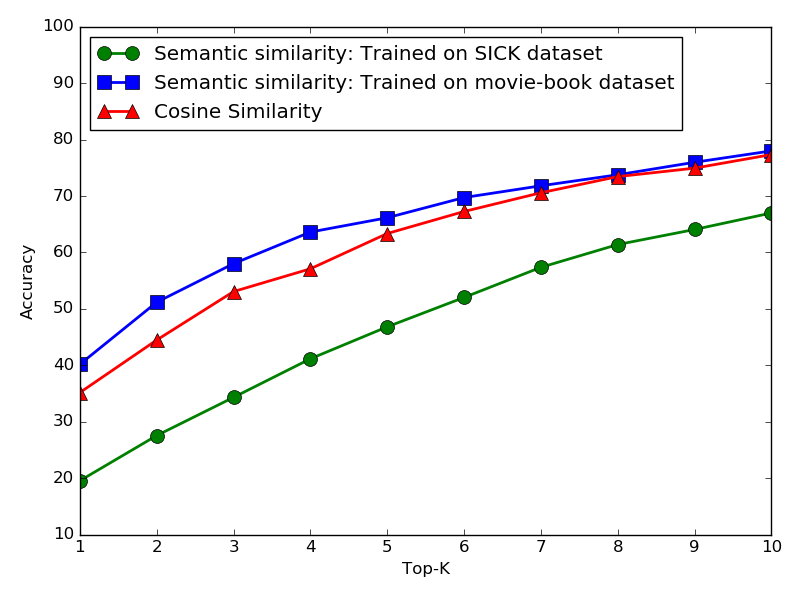}
	\caption{A cumulative match score curve (CMC) showing the rank-$1$ to rank-$10$ Movie Retrieval Accuracy (MRA) using the Dialog Model.}
	\label{fig:dialogk}
\end{figure}

\begin{figure}[!h]
	\centering
	\includegraphics[width=3.4in]{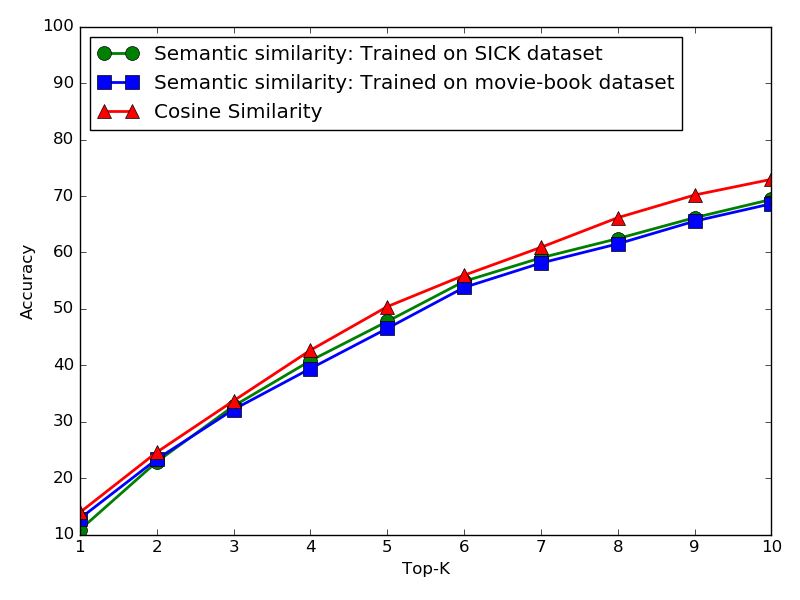}
	\caption{A cumulative match score curve (CMC) showing the rank-$1$ to rank-$10$ Movie Retrieval Accuracy (MRA) using the Visual Model.}
	\label{fig:visualk}
\end{figure}

The performance of the proposed pipeline is evaluated using top-$k$ Movie Retrieval Accuracy (MRA). This measure calculates the percentage of book sentences input for which all the retrieved movie clips are from the same movie as the input. The performance of the Dialog model and the Visual model are shown in Figure~\ref{fig:dialogk} and Figure~\ref{fig:visualk}, respectively. The major observations obtained from the results are as follows:
\begin{enumerate}
	\item A rank-$10$ MRA of $80\%$ is obtained for the Dialog model and $71\%$ MRA is obtained for the Visual model, using the proposed approach. The proposed semantic similarity fared better or comparable to the other two approaches, showing the effectiveness of the similarity method.
	\item Although the dataset provides the ground truth alignment, the exact aligned video snippet retrieval accuracy for a given book sentence is irrelevant for our experiments. For a given book sentence, there can be multiple parts in the movie that is semantically related and retrieving those movie snippets is the creative task at hand and not just the manually aligned movie snippet. Thus, the movie retrieval accuracy is a strong measure to evaluate our creative system rather than the exact alignment retrieval accuracy.
\end{enumerate}

\begin{figure}[!h]
	\centering
	\includegraphics[width=3.4in]{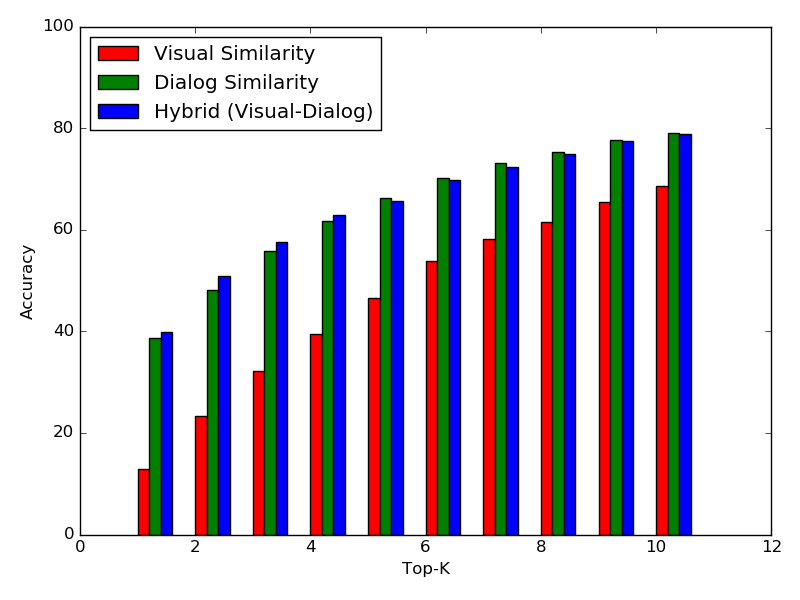}
	\caption{A cumulative match score curve (CMC) showing the rank-$1$ to rank-$10$ Movie Retrieval Accuracy (MRA) using the Hybrid Model.}
	\label{fig:hybridk}
\end{figure}

\begin{figure*}[ht]
	\centering
	\includegraphics[width=5.6in]{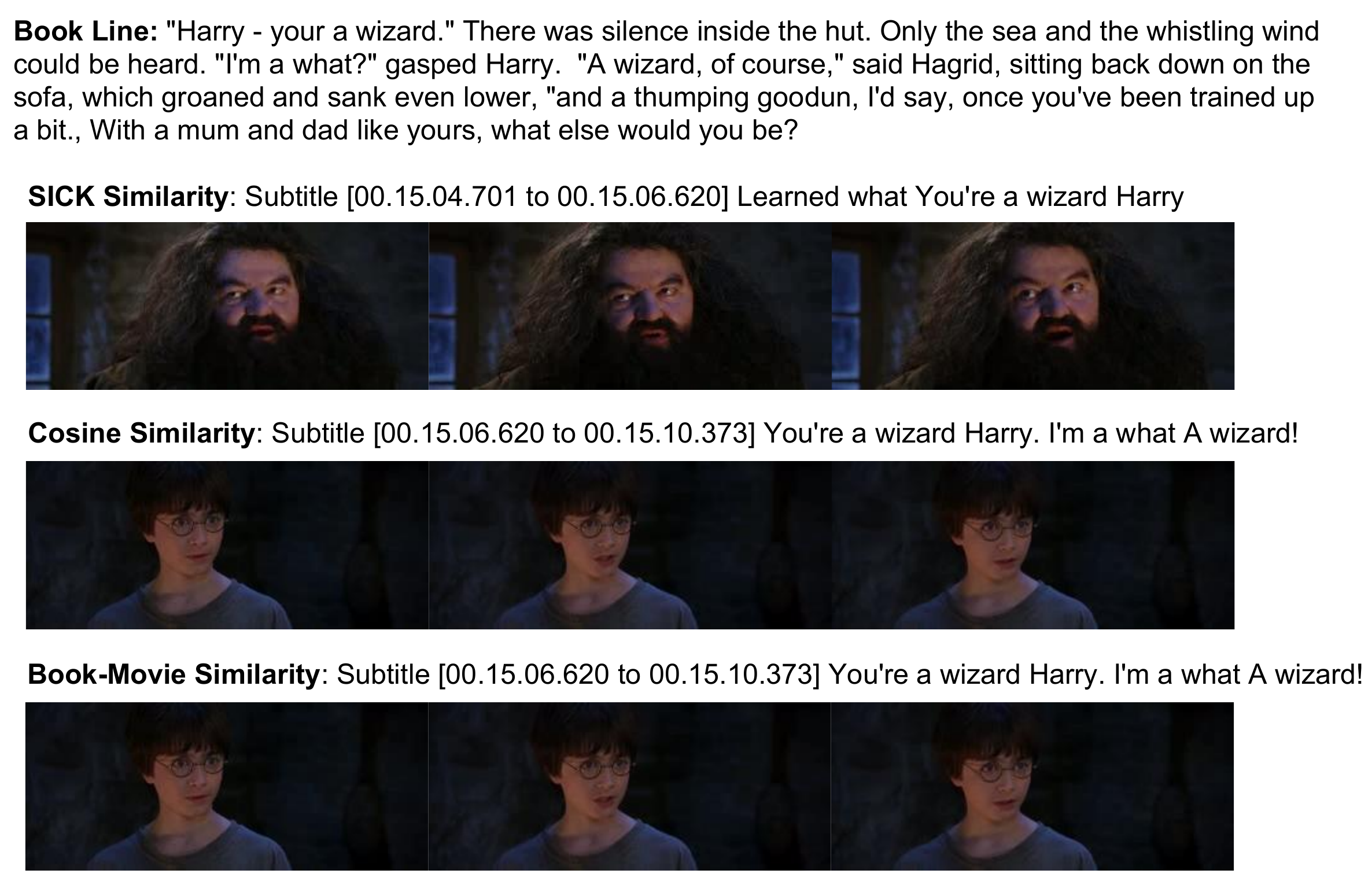}
	\caption{A working example of the results obtained by the Dialog Model. The input book line and the video retrievals from the ``Harry Potter: The Philosopher's Stone" movie using the three similarity measures.}
	\label{fig:hp}
\end{figure*}

\begin{figure*}[!h]
	\centering
	\includegraphics[width=5.6in]{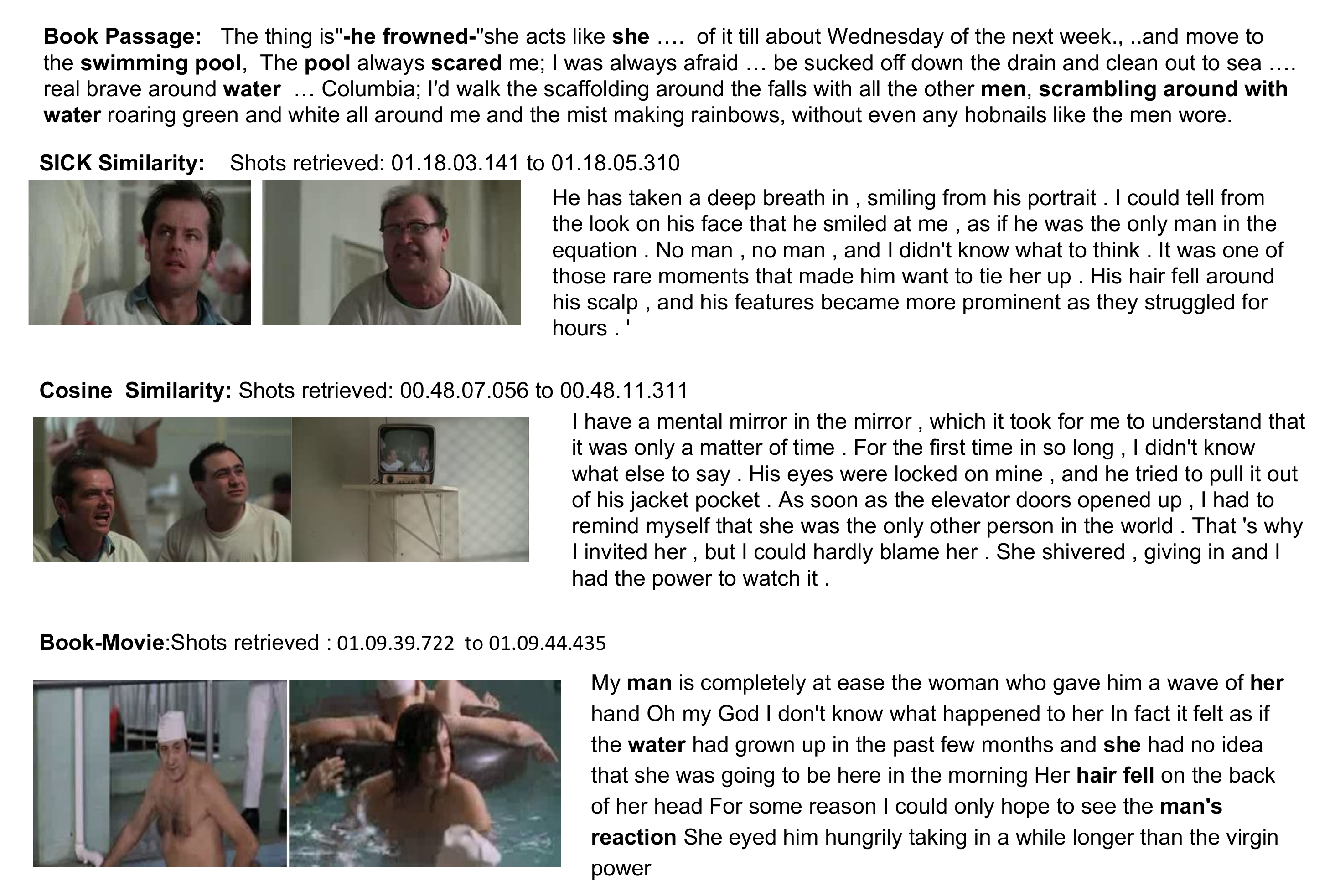}
	\caption{A working example of the results obtained by the Visual Model. The input book line and the video retrievals from the ``One Flew Over the Cuckoo's Nest" movie using the three similarity measures.}
	\label{fig:as}
\end{figure*}

To show the effectiveness of the combined Dialog and Visual Model, a Hybrid model was trained on the entire test set and the results are shown in Figure~\ref{fig:hybridk}. The results shows that the hybrid model performs better than the individual models at all ranks, suggesting to use both the modalities for matching during movie retrieval. From Figure~\ref{fig:hybridk}, it can observed that the Dialog model performs much better than the Visual model, suggesting that the dialog has richer information than the visual content. The same observation is extended to the Hybrid model, as the Hybrid does not show a rapid improvement compared to the Dialog model. However, there are certain caveats in this comparison as the Visual model is trained on a much smaller dataset compared with the Dialog model.
A working example of the Dialog Model and the Visual Model is shown in Figure~\ref{fig:hp} and Figure~\ref{fig:as}, respectively.

%% file: 7_Conclusion.tex
\section{Conclusion and Future Work}

In this research, we proposed a creative system which could visualize a snippet of book content using its corresponding movie visuals. We devised three models to retrieve semantically similar movie content of a book snippet: (i) a dialog model which use only the dialog content from the movie, (ii) a visual model which uses only the visual content from the movie, and (iii) a hybrid model which combines both the visual and dialog content from the movie. A frame-wise conditional LSTM based decoder is used to generate a single story explaining a movie snippet. Experimental results on the publicly available MovieBook dataset, shows the effectiveness of the proposed hybrid model providing around $80\%$ rank-$10$ retrieval accuracy.

In future, we plan to extend this approach by creatively generating animated images and video snippets that explains a book snippets~\cite{lin2015don}~\cite{vedantam2015learning}. Thus, the proposed pipeline could be used for unseen book or for books which do not have a corresponding movie and their corresponding visual abstractions could be generated. Such a creative system would eventually be of great use for creative directors and advertisment film makers as they can visualize stories and scripts before the movie is being produced.